\documentclass[11pt,a4paper]{article}

\usepackage[utf8]{inputenc}
\usepackage[T1]{fontenc}
\usepackage{lmodern}
\usepackage[margin=1in]{geometry}
\usepackage{booktabs}
\usepackage{tabularx}
\usepackage{longtable}
\usepackage{pdflscape}
\usepackage{graphicx}
\usepackage{hyperref}
\usepackage{xcolor}
\usepackage{listings}
\usepackage{amsmath,amssymb}
\usepackage{enumitem}
\usepackage{textcomp}
\usepackage{microtype}
\usepackage{authblk}

\hypersetup{
    colorlinks=true,
    linkcolor=blue!70!black,
    citecolor=blue!70!black,
    urlcolor=blue!70!black
}

\title{MedCalc-Bench Doesn't Measure What You Think:\\A Benchmark Audit and the Case for Open-Book Evaluation}
\author{Artus Krohn-Grimberghe}
\date{}

\begin{document}
\maketitle

\begin{abstract}
MedCalc-Bench is a widely used benchmark for evaluating LLM performance on clinical calculator tasks, with state-of-the-art direct prompting scores plateauing around 35\% on the Verified split (HELM MedHELM leaderboard) and the best published approach---RL with verifiable rewards---reaching 74\%. We present three contributions that challenge the benchmark's current framing. First, we conduct a systematic audit of the benchmark's calculator implementations, identifying and fixing over 20 errors ranging from critical formula inaccuracies to runtime bugs in a NeurIPS-published dataset. Second, we show that a simple intervention---providing the model with the calculator specification at inference time (``open-book'' prompting)---raises accuracy from ${\sim}52$\% to 81--85\% on GLM-4.6V and GLM-4.7, surpassing all published results including RL-trained systems, without any fine-tuning. Third, we establish an upper bound of 95--97\% using GPT-5.2-Thinking, with residual errors attributable primarily to ground-truth issues and dataset ambiguities. Our findings suggest that MedCalc-Bench predominantly measures formula memorization and arithmetic precision rather than clinical reasoning, and would be better framed as a tool-use evaluation.
\end{abstract}

\section{Introduction}

Clinical calculators---standardized scoring systems such as CKD-EPI, APACHE-II, MELD, and the Glasgow Coma Scale---are routine instruments in hospital medicine. They transform patient observations into risk scores, dosage adjustments, or staging classifications. MedCalc-Bench (NeurIPS 2024 Datasets Track) operationalizes 55 such calculators as an LLM evaluation task: given a clinical vignette, the model must identify the relevant calculator, extract the required parameters, and compute the correct result.

On the Stanford HELM MedHELM leaderboard, the best direct-prompting results reach only 34.8\% accuracy on MedCalc-Bench Verified. Specialized approaches---RL with verifiable rewards (MedCalc-R1), agentic tool frameworks (RiskAgent), and extended thinking with code execution (Claude Opus 4.5)---push this to 61--74\%, but the task remains among the hardest in clinical NLP benchmarking.

We start from a simple observation: in clinical practice, no physician memorizes the APACHE-II formula. They use a calculator. The benchmark's difficulty may therefore stem less from clinical reasoning and more from an artificial constraint---requiring the model to recall, from parametric memory, the exact formula and execute multi-step arithmetic on real-valued inputs (logarithms, exponents) where even minor precision errors cascade outside the benchmark's acceptance band.

This paper makes three contributions:

\begin{enumerate}
    \item \textbf{Benchmark audit.} We identify and document over 20 errors in the official calculator implementations, including incorrect formula coefficients (CKD-EPI, MELD-Na, Child-Pugh), missing scoring criteria (SIRS, SOFA), key typos that silently zero out score components (CCI, Cardiac Risk Index), and broken file paths. Several of these errors affect ground-truth labels in the evaluation set.

    \item \textbf{Open-book prompting.} We propose a minimal intervention: appending the calculator's specification (formula, parameter definitions, scoring rules) to the prompt. With no fine-tuning, no retrieval augmentation, and optional code execution, this raises GLM-4.6V from 52\% to 81.5\% and GLM-4.7 from 36\% (vanilla) to 85.5\% on a 275-row subsample---exceeding all published results.

    \item \textbf{Upper-bound analysis with GPT-5.2-Thinking.} We run GPT-5.2-Thinking (xhigh reasoning effort, with search and tools) on the 198 rows that GLM-4.6V fails under open-book prompting. It resolves 140/198, with a subsequent audit attributing 19 of the remaining 58 errors to ground-truth or implementation issues and 10 to genuine ambiguity. This yields a conservative upper-bound accuracy of 94.7\% and an optimistic bound of 97.4\%.
\end{enumerate}

Taken together, these results suggest that MedCalc-Bench, as currently constituted, is primarily a test of formula recall and decimal arithmetic under tight tolerance conditions ($\pm$5\% for equation-based; exact match for rule-based)---not of clinical reasoning. We argue it should be repositioned as a tool-use benchmark, where the evaluation focuses on parameter extraction and appropriate calculator and unit converter selection rather than formula memorization.

\section{Background and Related Work}

\subsection{MedCalc-Bench}
\label{sec:medcalcbench}

MedCalc-Bench was introduced at NeurIPS 2024 as a dataset of clinical calculator tasks spanning 55 calculators in two categories: equation-based (continuous-valued outputs evaluated within $\pm$5\% tolerance) and rule-based (exact categorical match). MedCalc-Bench Verified is the maintained successor with corrected labels. The benchmark provides three evaluation settings: zero-shot, one-shot, and code-augmented.

\subsection{Consolidated Leaderboard}

Table~\ref{tab:leaderboard} synthesizes all published MedCalc-Bench results we could identify through a structured literature survey using seven different research-capable LLM systems (OpenAI Deep Research, Gemini Deep Research, Kimi K2 Deep Research, Qwen~3, Grok~4 Expert, GPT-5.2-Pro). Results were consolidated and cross-validated by GPT-5.2-Pro with reference verification by GPT-5.2-Thinking and Claude Opus~4.5.

\begin{landscape}
\begin{table}[p]
\centering
\caption{Published MedCalc-Bench results, sorted by accuracy descending. Results span different dataset variants (Orig, Verified, Cleaned) and are not directly comparable; see Section~\ref{sec:medcalcbench} for variant descriptions.}
\label{tab:leaderboard}
\small
\begin{tabular}{@{}clllllc@{}}
\toprule
Rank & Approach & Model/System & Dataset & Setting & Accuracy & Src \\
\midrule
1  & RL + verifiable rewards    & DeepSeek-R1 (MedCalc-R1)           & Orig (train/test 9,765/1,048)          & SFT + GRPO                & 73.95\%               & \cite{medcalcr1} \\
2  & RL on recomputed labels    & Qwen3-8B                           & Orig (recomputed; test $n$=1,047)      & GRPO                      & 71.4\%                & \cite{scalably} \\
3  & RL + verifiable rewards    & o1-mini                            & Orig (train/test 9,765/1,048)          & SFT + GRPO                & 67.84\%               & \cite{medcalcr1} \\
4  & Agentic tool framework     & RiskAgent-GPT-4o                   & Orig (external eval)                   & Zero-shot CoT + tools     & 67.71\%               & \cite{riskagent} \\
5  & Extended thinking + tools  & Claude Opus 4.5                    & Bench (company eval)                   & 64k thinking + Python     & 61.3\%                & \cite{anthropic} \\
6  & Step-wise eval diagnostic  & GPT-4o                             & Cleaned \& restructured                & Step-wise final-answer    & 62.7\%$\to$43.6\%     & \cite{scores2steps} \\
7  & Agentic + step-wise eval   & MedRaC                             & Cleaned \& restructured                & Retrieval + Python        & up to 53.19\%         & \cite{scores2steps} \\
8  & RL + verifiable rewards    & MedCalc-R1 (3B)                    & Orig (train/test 9,765/1,048)          & SFT + GRPO                & 51.34\%               & \cite{medcalcr1} \\
9  & Prompting baseline         & GPT-4                              & Orig (test 1,047)                      & One-shot CoT              & 50.9\%                & \cite{medcalcbench} \\
10 & Supervised fine-tuning     & Mistral-7B                         & Orig (train split)                     & SFT                       & 49.19\%               & \cite{neurips_supp} \\
11 & Code-exec prompting        & GPT-4                              & Orig                                   & Zero-shot CoT + code      & 48.51\%               & \cite{neurips_supp} \\
12 & Supervised fine-tuning     & Llama-2-7B                         & Orig (train split)                     & SFT                       & 45.75\%               & \cite{neurips_supp} \\
13 & Direct prompting (HELM)    & DeepSeek-R1 / Gemini 2.5 Pro       & Verified                               & Zero-shot                 & 34.8\%                & \cite{helm} \\
14 & Code-exec prompting        & GPT-3.5                            & Orig                                   & Zero-shot CoT + code      & 30.29\%               & \cite{neurips_supp} \\
\bottomrule
\end{tabular}
\end{table}
\end{landscape}

\subsection{Why Is This Benchmark Hard?}

The low scores invite decomposition. The task requires three capabilities in sequence:

\begin{enumerate}
    \item \textbf{Parameter extraction}---identifying the correct clinical values from a vignette, anchored to the right admission date when multiple encounters are described.
    \item \textbf{Formula recall}---retrieving the exact scoring rules or equations from parametric memory, including version-specific details (e.g., MELD vs.\ MELD~3.0, CKD-EPI 2009 vs.\ 2021).
    \item \textbf{Arithmetic execution}---computing results involving unit conversions, logarithms, exponents, and real-valued coefficients to within the benchmark's $\pm$5\% tolerance band.
\end{enumerate}

Failure at any step cascades to an incorrect final answer. Prior work has focused on improving steps 2 and 3 through RL training or code execution. We focus on the simpler question: what happens if you eliminate step~2 entirely?

\section{Benchmark Audit}

\subsection{Methodology}

We conducted a systematic review of all 55 calculator implementations in the MedCalc-Bench repository. Each implementation was cross-referenced against its primary clinical source (original publication, MDCalc documentation, or clinical guideline). Test suites were written for each calculator. Errors were identified through a combination of automated testing, manual review, and LLM-assisted code analysis (GPT-5.2-Thinking, GPT-5.2-Pro, Claude Opus~4.5).

\subsection{Error Taxonomy}

We identified and fixed over 20 errors across four categories. Table~\ref{tab:errors} summarizes the findings.

\begin{table}[htbp]
\centering
\caption{Calculator implementation errors by category.}
\label{tab:errors}
\small
\begin{tabularx}{\textwidth}{@{}lrX@{}}
\toprule
Category & Count & Examples \\
\midrule
Logic \& formula errors & 9 & CKD-EPI 2021 male coefficient ($0.7 \to 0.9$); MELD-Na updated to MELD~3.0; HOMA-IR unit conversion ($\times \to \div$); Child-Pugh bilirubin MW ($548.66 \to 584.66$); MME fentanyl patch factor ($0.13 \to 2.4$ per CDC 2022); HEART score atherosclerotic disease logic; SIRS missing band neutrophil criterion; SOFA FiO2 handling and vasopressor ranges; GCS ``Not Testable'' handling \\
\addlinespace
Runtime \& implementation bugs & 5 & Framingham argument order mismatch in unit converter; CCI key typo (\texttt{liver\_diease}); Cardiac Risk Index key typo (\texttt{ischemetic}); broken file paths for MDRD and CURB-65 \\
\addlinespace
Threshold \& boundary errors & 3 & Glasgow-Blatchford BUN=70 boundary ($4 \to 6$~pts); CURB-65 BUN threshold ($>$19 $\to$ $>$20~mg/dL); FIB-4 platelet scaling \\
\addlinespace
Precision \& rounding errors & 3 & Steroid conversion intermediate rounding; significant digits utility off-by-one; QTc Framingham formula display ($\div \to +$) \\
\bottomrule
\end{tabularx}
\end{table}

\subsection{Impact on Ground Truth}

Several of these errors propagate to ground-truth labels. For example, the CCI key typo (\texttt{liver\_diease} $\to$ \texttt{liver\_disease}) silently zeros out liver disease points for every patient in the dataset with liver disease, producing systematically incorrect reference scores. The APACHE-II implementation has a mapping mismatch (\texttt{severe\_organ\_failure\_or\_immunocompromise} vs.\ \texttt{organ\_failure\_immunocompromise}) that prevents chronic health points from being applied. These affect entire calculator categories.

We will report these issues upstream upon publication.

\section{Method: Open-Book Prompting}
\label{sec:method}

\subsection{Motivation}

In clinical practice, calculators are tools. A physician using the CKD-EPI equation does not derive it from memory; he enters lab values into a validated calculator. MedCalc-Bench's default framing---requiring the model to both recall the formula and execute it---conflates tool knowledge with tool use.

We propose a minimal change that we term \emph{open-book prompting}: append the calculator specification to the prompt. The specifications used in our experiments are those provided by the MedCalc-Bench repository itself (the Python implementations and their docstrings). We additionally prepared a revised set of specifications with corrected formulas and explicit online references for each calculator, which we will release alongside this paper. The specification includes the formula (or scoring table), parameter definitions with expected units, and any version-specific notes. The model's task reduces to:

\begin{enumerate}
    \item Identifying the correct reference encounter from the vignette.
    \item Identify and extract the required parameter values from the vignette.
    \item Apply the provided formula or scoring rules.\footnote{We actually forgot to supply unit conversion guidance to the model but realized this only at the very last step of our experiments once we looked into the instances where GPT-5.2 answered incorrectly.}
    \item Return the computed result.
\end{enumerate}

This is strictly less information than what a code-augmented approach provides (which gives the model executable code), but strictly more than zero-shot or one-shot prompting (which provides no formula).

\subsection{Prompt Design}

We evaluated three prompt variants:

\begin{itemize}[nosep]
    \item \textbf{Baseline (vanilla).} The original MedCalc-Bench Verified prompt template, equivalent to the ``mistral'' setting in the evaluation code. No formula provided.
    \item \textbf{Open-book.} The baseline prompt plus the full calculator specification (formula, parameter definitions, scoring rules, unit expectations).
    \item \textbf{Open-book + structured guidance.} The open-book prompt plus explicit extraction and computation steps (``scratchpad'' chain-of-thought).
\end{itemize}

See Appendix~\ref{app:prompts} for select examples.

\subsection{Models}

We selected models primarily for cost efficiency and availability rather than peak capability, to demonstrate that the intervention---not the model---drives the improvement:

\begin{itemize}[nosep]
    \item \textbf{GLM-4.6V} (Zhipu AI): Mid-range multimodal model. Primary evaluation model, run on all 1,100 test rows.
    \item \textbf{GLM-4.7} (Zhipu AI): Updated variant. Run on a 275-row subsample for comparison.
    \item \textbf{GPT-5.2-Thinking} (OpenAI, ``xhigh'' reasoning effort): Used only for upper-bound analysis on residual errors, not as a primary evaluation model.
\end{itemize}

\section{Experiments}

\subsection{Setup}

All experiments use the MedCalc-Bench Verified test split (1,100 rows, 55 calculators). Evaluation follows the benchmark's standard protocol: exact match for rule-based calculators, $\pm$5\% tolerance band (Lower/Upper limits) for equation-based calculators. We use our corrected calculator implementations (Section~3) for ground-truth comparison where applicable.

For GLM-4.7, we evaluated a stratified 275-row subsample (fixed proportional representation across all 55 calculators) due to time constraints.

We note that our vanilla baseline for GLM-4.6V (51.9\%) substantially exceeds the HELM direct-prompting results (34.8\% for DeepSeek-R1 / Gemini~2.5~Pro). Both GLM models had access to a Python execution environment, though we did not explicitly instruct them to use it. The two models were consumed through different harnesses for these evaluation runs. This likely explains why GLM-4.6V outperforms its successor in the vanilla setting---GLM-4.6V may have silently offloaded arithmetic to Python on some fraction of rows, making its 51.9\% baseline a tool-``contaminated'' result rather than a strictly vanilla one. This interpretation is consistent with GLM-4.7's vanilla score (36.0\%) falling in the ballpark of the best reported direct-prompting results on HELM (34.8\%). We report GLM-4.6V's numbers as obtained, but the distinction does not materially affect our central finding: the open-book intervention dominates regardless of whether the baseline includes incidental tool use.

\subsection{Main Results}

\begin{table}[htbp]
\centering
\caption{Accuracy by prompt condition.}
\label{tab:results}
\begin{tabular}{@{}llrrr@{}}
\toprule
Model    & Prompt               & $N$   & Correct & Accuracy \\
\midrule
GLM-4.6V & Vanilla (baseline)   & 1,100 & 571     & 51.9\% \\
GLM-4.7  & Vanilla (baseline)   & 275   & 99      & 36.0\%\textsuperscript{$\dagger$} \\
GLM-4.6V & Open-book            & 1,100 & 896     & \textbf{81.5\%} \\
GLM-4.7  & Open-book            & 275   & 235     & \textbf{85.5\%} \\
GLM-4.6V & Open-book + guidance & 1,100 & 902     & \textbf{82.0\%} \\
GLM-4.7  & Open-book + guidance & 275   & 231     & 84.0\% \\
\bottomrule
\multicolumn{5}{@{}l}{\footnotesize \textsuperscript{$\dagger$} It is unclear why GLM-4.7 performs so much worse than its predecessor GLM-4.6V} \\
\multicolumn{5}{@{}l}{\footnotesize in the vanilla setting. We did not further analyze this discrepancy because our focus is} \\
\multicolumn{5}{@{}l}{\footnotesize the relative effect of open-book prompting.} \\
\end{tabular}
\end{table}

The open-book prompt yields a \textbf{+29.6 percentage point improvement} for GLM-4.6V (51.9\% $\to$ 81.5\%), surpassing all published results including RL-trained systems (DeepSeek-R1 MedCalc-R1: 74.0\%) and agentic frameworks with code execution (Claude Opus~4.5: 61.3\%; RiskAgent: 67.7\%).

The additional structured guidance provides marginal benefit (+0.5~pp for GLM-4.6V), suggesting that the formula specification itself---not the chain-of-thought scaffolding---accounts for nearly all of the gain.

\subsection{Error Analysis: Baseline}

Under the vanilla prompt, we observe three dominant failure modes:

\begin{enumerate}
    \item \textbf{Arithmetic errors on real-valued operations.} Exponentiations and logarithms of decimal values are frequently slightly off, placing results outside the $\pm$5\% acceptance band. This is a known limitation of autoregressive token prediction for numerical computation.
    \item \textbf{Wrong reference date.} When vignettes describe multiple encounters, models sometimes extract values from the wrong admission, leading to parameter errors upstream of any computation.
    \item \textbf{Formula version conflicts.} Multiple valid versions exist for several calculators (e.g., MELD vs.\ MELD~3.0, CKD-EPI 2009 vs.\ 2021). The model may recall a different version than the one the benchmark expects.
\end{enumerate}

\subsection{Error Analysis: Open-Book}

Under the open-book prompts, the residual errors on GLM-4.6V shift to a different profile---predominantly parameter extraction failures and arithmetic precision issues, with formula recall errors effectively eliminated. The best-performing variant (open-book + structured guidance) yields 198 residual errors (18.0\%), which form the basis for the upper-bound analysis in Section~6.

\section{Upper-Bound Analysis with GPT-5.2-Thinking}

\subsection{Protocol}

To estimate how much of the residual error is attributable to model limitations versus dataset issues, we ran GPT-5.2-Thinking (reasoning effort: xhigh, with search and tools enabled) on the 198 rows where GLM-4.6V (open-book + structured guidance) produced incorrect answers. GPT-5.2-Thinking received the same open-book prompt.

\subsection{Results}

GPT-5.2-Thinking resolved 140 of 198 residual errors (70.7\% recovery rate). For the remaining 58 errors, we conducted a manual audit informed by GPT-5.2-Thinking's self-assessment:

\begin{table}[htbp]
\centering
\caption{GPT-5.2-Thinking residual error attribution ($N$=58).}
\label{tab:residual}
\begin{tabular}{@{}lrl@{}}
\toprule
Category & Count & Description \\
\midrule
Ground-truth or evaluation issue & 19 & APACHE-II mapping mismatch (10 rows), \\
                                 &    & unit conflicts between vignette and GT (4 rows), \\
                                 &    & other labeling inconsistencies \\
Likely model error               & 29 & Genuine failures on parameter extraction \\
                                 &    & or computation \\
Ambiguous                        & 10 & Insufficient information to adjudicate \\
\bottomrule
\end{tabular}
\end{table}

\subsection{Composite Accuracy Estimate}

Assuming GPT-5.2-Thinking would correctly solve all rows that the weaker GLM-4.6V solves (a reasonable assumption given capability ordering):

\begin{itemize}[nosep]
    \item \textbf{Conservative upper bound:} $(902 + 140) / 1{,}100 = \mathbf{94.7\%}$
    \item \textbf{Optimistic upper bound:} $(902 + 140 + 19 + 10) / 1{,}100 = \mathbf{97.4\%}$
\end{itemize}

The true figure potentially falls closer to the optimistic bound, as the 19 ground-truth issues were corroborated by implementation bugs documented in Section~3.

\section{Discussion}

\subsection{What Does MedCalc-Bench Actually Measure?}

Our results decompose MedCalc-Bench difficulty into three components:

\begin{enumerate}
    \item \textbf{Formula recall} (${\sim}$30~pp). The difference between vanilla and open-book prompting. This is the largest single factor and is trivially eliminated by providing the specification.
    \item \textbf{Arithmetic precision} (${\sim}$5--13~pp). The gap between open-book mid-range models and GPT-5.2-Thinking. Partially addressable through code execution.
    \item \textbf{Parameter extraction} (${\sim}$3--5~pp). The irreducible core of the task---correctly parsing clinical vignettes. This is the component most relevant to clinical reasoning.
\end{enumerate}

The benchmark, as currently scored, is dominated by components 1 and 2, which are not meaningful tests of clinical capability. A physician who cannot recall the APACHE-II formula is not incompetent; a physician who extracts the wrong lab value from a chart is.

\subsection{Recommendation: Reframe as Tool-Use Benchmark}

We propose that MedCalc-Bench (or a successor) adopt an open-book, tool-enabled setting as the default evaluation, with the task explicitly framed as:

\begin{itemize}[nosep]
    \item \textbf{Input:} Clinical vignette + calculator specification (formula, parameters, units, conversions).
    \item \textbf{Output:} Extracted parameter values + computed result.
    \item \textbf{Evaluation:} Score parameter extraction and final answer separately.
\end{itemize}

This reframing would make the benchmark a more valid measure of clinical NLP capability and would align with how calculators are actually used in practice.

\subsection{AI-Assisted Research Workflow}

This work was conducted by a single researcher using a multi-model workflow, with different LLMs selected for distinct roles based on cost, capability, and availability trade-offs. We document the allocation explicitly, both for transparency and because the workflow itself illustrates a broader point about how frontier AI tools reshape individual research productivity.

\begin{landscape}
\begin{table}[p]
\centering
\caption{Model roles and selection rationale.}
\label{tab:workflow}
\small
\begin{tabularx}{\linewidth}{@{}lp{4.5cm}X@{}}
\toprule
Role & Model(s) & Rationale \\
\midrule
Primary evaluation (1,100 rows) & GLM-4.6V, GLM-4.7 & Low inference cost; sufficient capability to demonstrate the prompt intervention effect. Deliberately mid-range to show that not the model but the method drives the improvement. \\
\addlinespace
Literature survey & OpenAI Deep Research, Gemini Deep Research, Kimi K2 Deep Research, Qwen~3, Grok~4 Expert & Six parallel deep-research runs to maximize coverage of published MedCalc-Bench results. Extended by GPT-5.2-Pro and cross-validated by GPT-5.2-Thinking for consistency. \\
\addlinespace
Leaderboard consolidation \& medical/code review & GPT-5.2-Pro & By far the strongest publicly available reviewer model for cross-checking extracted results, validating calculator implementations against clinical sources, and auditing ground-truth labels. Used via ChatGPT to avoid API cost. \\
\addlinespace
Code generation, test suites, automation & GPT-5.2 via Codex & All calculator fixes, test suite authoring, evaluation pipeline code, and multi-hour automation tasks (up to 224 minutes continuous operation). \\
\addlinespace
Upper-bound analysis (198 residual rows) & GPT-5.2-Thinking (xhigh) & Maximum reasoning capability for difficult residual cases. Search and tools enabled. \\
\addlinespace
Reference validation; additional sparring & Claude Opus 4.5 & Validated leaderboard references; served as a second, independent opinion to GPT-5.2-Pro. \\
\addlinespace
Writing & Claude Opus 4.6 & Claude Opus has a great text and task understanding. Opus' text is usually received very well and the model allows for quick iterations compared to GPT-5.2-Pro, which can also write well but with a very surgical tone. \\
\bottomrule
\end{tabularx}
\end{table}
\end{landscape}

The total API spend beyond existing subscription costs was negligible---a deliberate constraint. The primary evaluation models (GLM-4.6V/4.7) were chosen specifically because they are inexpensive, making the results reproducible without significant compute budgets. The frontier models (GPT-5.2 family, Claude Opus~4.5) were accessed through consumer chat and coding tool subscriptions rather than API billing.

It is worth stating plainly: this project would not have been feasible as a solo effort without the late-2025/early-2026 generation of frontier models. GPT-5.2 was the primary workhorse across reasoning, code, and---deserving particular credit---web search, where its ability to locate, cross-reference, and synthesize clinical literature, benchmark papers, and implementation details was transformative. The literature survey that produced Table~\ref{tab:leaderboard} would have taken weeks of manual work; GPT-5.2's search capabilities compressed it into hours. GPT-5.2 via Codex sustained multi-hour autonomous coding sessions (up to 224~minutes) for calculator fixes, test suites, and evaluation pipeline work. GPT-5.2-Thinking's sharp reasoning was essential for the upper-bound analysis, ground-truth auditing, and resolving genuinely ambiguous clinical vignettes.

The Chinese model ecosystem provided distinct and sometimes unique contributions. Qwen3, GLM-4.7, and Kimi K2---for this task all accessed via free web chat---surfaced Chinese-language references and clinical perspectives that Western-centric models missed, and occasionally produced insights no other model offered as readily. Grok~4 Expert contributed to a parallel survey. GLM-4.6V and GLM-4.7 served as cost-effective evaluation workhorses, demonstrating that the method generalizes beyond frontier models.

Claude Opus~4.5 served a different but essential role: as a sparring and reflection partner. Where GPT-5.2-Pro was the dominant analytical engine, Opus~4.5 provided a genuinely independent second perspective---catching hallucinated references that multiple other models had introduced during the literature survey, stress-testing claims, and helping steer the project's narrative and structure. The complementarity between these two frontier models---one for deep analytical work, the other for critical review and reflection---proved more valuable than either alone.

The step change in capability between this model generation and its predecessors is what makes ``4~weekends, one researcher'' a viable production mode for benchmark-scale audits.

This workflow---cheap models for bulk evaluation, frontier models for review and hard cases, specialized tools for code---may generalize as a template for resource-constrained research projects leveraging current AI capabilities.

\subsection{Limitations}

\begin{itemize}[nosep]
    \item \textbf{Model selection.} Our primary evaluation uses GLM-4.6V and GLM-4.7, which are mid-range models. Results on frontier models (GPT-5.2, Claude Opus~4.5, Gemini~2.5~Pro) would strengthen the findings but were not run due to cost constraints.
    \item \textbf{Dataset version.} Our experiments use the MedCalc-Bench Verified test split (1,100 rows) with our corrected calculator implementations. Cross-variant comparisons with results reported on the original MedCalc-Bench should be interpreted with caution.
    \item \textbf{Capability monotonicity assumption.} The composite accuracy estimate assumes GPT-5.2-Thinking $\geq$ GLM-4.6V on all rows, which may not hold in every case.
    \item \textbf{Sample size for GLM-4.7.} The 275-row subsample for GLM-4.7 limits the precision of comparative claims.
\end{itemize}

\section{Conclusion}

MedCalc-Bench is hard---but largely for the wrong reasons. The benchmark's difficulty is dominated by formula memorization and arithmetic precision, not by the clinical reasoning it purports to measure. A simple prompt-level intervention---providing the calculator specification (open-book prompting)---outperforms RL-trained models, agentic frameworks, and extended-thinking systems, at a fraction of the compute cost. Meanwhile, the benchmark's own calculator implementations contain over 20 errors that affect ground-truth labels. We release our corrected implementations, consolidated leaderboard, and prompt templates to support future work.


\appendix

\section{Prompt Templates}
\label{app:prompts}

\subsection{Original Prompt Template (Baseline)}

\begin{lstlisting}
<s>[INST] You are a helpful assistant for calculating a score for a given
patient note. Please think step-by-step to solve the question and then
generate the required score. Your output should only contain a JSON dict
formatted as {"step_by_step_thinking":
str(your_step_by_step_thinking_procress_to_solve_the_question), "answer":
str(short_and_direct_answer_of_the_question)}.
Here is the patient note:
{{ note }}

Here is the task:
{{ question }}

Please directly output the JSON dict formatted as
{"step_by_step_thinking":
str(your_step_by_step_thinking_procress_to_solve_the_question), "answer":
str(short_and_direct_answer_of_the_question)}: [/INST]
\end{lstlisting}

\subsection{Common Failure Modes Under the Original Prompt}

\subsubsection{Incorrect reference encounter selection}

The benchmark's vignettes often contain multiple encounters (outpatient visits, ED presentations, inpatient admissions). The correct rule is to extract values from the index encounter---typically the first ED or inpatient admission---but models frequently select values from the wrong encounter.

\textbf{Example.} A vignette describes a 51-year-old male with Creatinine 1.0~mg/dL at an outpatient visit on 28th July 2016, and Creatinine 2.0~mg/dL at an ED admission on 3rd August 2016. GLM-4.7 selected the outpatient value (1.0), producing an incorrect Cockcroft-Gault result. When prompted ``When was the patient first admitted?'', the model confirmed:

\begin{quote}
The patient was first admitted to the cardiology outpatient clinic on \textbf{28th July 2016}. (Note: This is described as an outpatient presentation; he was later admitted to the emergency department on 3rd August 2016).
\end{quote}

The model correctly identified both encounters but incorrectly classified the outpatient visit as an ``admission.'' The ED value (2.0~mg/dL) is the correct one for the index encounter.

This failure motivated the encounter selection rules added to the structured guidance prompt (see Section~\ref{app:structured}).

\subsubsection{Arithmetic precision on real-valued operations}

Models frequently produce slightly incorrect results for exponentiations and logarithms of decimal values:

\textbf{Example.} GLM-4.7 without tools computed $0.9938^{53} = 0.724583$, while the correct value is $0.7192$. This ${\sim}$0.7\% error is enough to place the final result outside the benchmark's $\pm$5\% acceptance band when compounded with other intermediate calculations.

\subsection{Open-Book Prompt Template}

\begin{lstlisting}
System:
You are a clinical extraction assistant. Use only the patient note. The
calculator specification below is authoritative for parameters, units,
conversions, and formula. Extract parameter values in canonical units. For
each parameter, return an object with "value" and "unit". If a parameter is
missing, set its value to null and note it in extraction_notes. If you can
compute the final value, compute it; otherwise, set answer to "N/A". Output
only a JSON object with the exact schema: {"parameters": {...},
"extraction_notes": [...], "answer": "<value>"} and nothing else. Keep
"answer" as the last field in the JSON.

User:
Patient note:
{{ note }}

Calculator specification:
{{ calculator_spec }}

Task:
{{ question }}

Return only the JSON object.
\end{lstlisting}

\subsection{Structured Guidance Extensions}
\label{app:structured}

The following encounter selection and workflow rules were appended to the open-book prompt for the structured guidance variant:

\begin{lstlisting}
CRITICAL ENCOUNTER SELECTION RULES:
1) The patient note may contain multiple encounters (outpatient visits, ED
   visits, admissions, follow-ups).
2) INDEX ENCOUNTER (the only encounter you may use for parameter values) is
   defined as: the first ED or inpatient admission described in the note.
   Identify it by setting keywords such as: "emergency department", "ED",
   "admitted", "hospitalized".
3) Do NOT use values from outpatient baseline visits or later follow-ups
   unless the index encounter is missing a required parameter.
4) If a required parameter is missing in the index encounter, use the
   closest earlier value in time as a fallback and explicitly note that
   fallback.

WORKFLOW (must follow in this order):
A) Build a timeline: list each encounter with date (if present) and setting
   (outpatient/ED/inpatient/follow-up).
B) Declare the selected INDEX ENCOUNTER (date + setting) based on the rules
   above.
C) Extract parameter values ONLY from the index encounter (or explicit
   fallback if missing).
D) Apply unit conversions to canonical units.
E) Apply weight-selection logic as specified.
F) Compute the final answer if possible.

CONFLICT HANDLING: If the note contains multiple candidate values for the
same parameter, you must (i) list the alternatives, (ii) state which one
belongs to the index encounter, and (iii) justify the choice.
\end{lstlisting}

\textbf{Example output under structured guidance} (same vignette as above):

\begin{lstlisting}
{
  "parameters": {
    "age": 51,
    "creatinine": 2.0,
    "height": 1.57,
    "sex": "Male",
    "weight": 49.0
  },
  "extraction_notes": [
    "Timeline:",
    "1. 28th July 2016: Outpatient clinic. Creatinine 1.0 mg/dL.",
    "2. 3rd August 2016: Emergency Department (Index Encounter). Creatinine 2.0 mg/dL.",
    "3. 12th August 2016: Admission for surgery.",
    "4. 28th February 2017: Follow-up.",
    "",
    "Index Encounter: 3rd August 2016 (Emergency Department).",
    "Parameter Extraction (from Index Encounter):",
    "- Age: 51 years.",
    "- Sex: Male (gentleman).",
    "- Height: 157 cm (1.57 m) from 28th July visit (fallback).",
    "- Weight: 49 kg from 28th July visit (fallback).",
    "- Creatinine: 2.0 mg/dL (index encounter on 3rd August).",
    "",
    "Calculations:",
    "- BMI: 49 / (1.57)^2 = 19.88 kg/m^2 (Normal range).",
    "- IBW (Devine, Male): 50 + 2.3*(61.81 - 60) = 54.16 kg.",
    "- Selected Weight: min(49.0, 54.16) = 49.0 kg.",
    "- CrCl: ((140 - 51) * 49.0) / (72 * 2.0) = 30.28 mL/min."
  ],
  "answer": "30.28"
}
\end{lstlisting}

\end{document}